\documentclass{article}
\usepackage{spconf,amsmath,graphicx}
\usepackage{subcaption}
\usepackage{booktabs}
\usepackage{array}
\usepackage{xspace}
\usepackage{cases}
\usepackage{multirow}
\usepackage{url}
\usepackage{xcolor}
\usepackage{paralist}
\usepackage{enumitem}
\usepackage[lined,ruled,linesnumbered]{algorithm2e}
\usepackage{algorithmic}
\usepackage{amsmath}
\usepackage{mathtools}
\usepackage{amssymb}
\usepackage[export]{adjustbox}

\graphicspath{ {Figures/} }
\usepackage{etoolbox}\AtBeginEnvironment{algorithmic}{\small}
\newif\ifsubmit

\submittrue

\ifsubmit
\newcommand{\ycwang}[1]{}

\newcommand{\sychien}[1]{}

\newcommand{\wctu}[1]{}

\newcommand{\cwwu}[1]{}

\newcommand{\ctliu}[1]{}

\else
\newcommand{\ycwang}[1]{{\bf \textcolor{cyan}{Y.-C. Wang: #1}}}

\newcommand{\sychien}[1]{{\bf \textcolor{purple}{S.-Y. Chien: #1}}}

\newcommand{\wctu}[1]{{\bf \textcolor{magenta}{W.-C. Tu: #1}}}

\newcommand{\cwwu}[1]{{\bf \textcolor{blue}{C.-W. Wu: #1}}}

\newcommand{\ctliu}[1]{{\bf \textcolor{brown}{C.-T. Liu: #1}}}

\fi

\newcommand{\mycaption}[2]{\caption{\textbf{#1.}~#2}}
\newcommand{\Lagr}{\mathcal{L}}
\usepackage{pifont}
\makeatletter
\DeclareRobustCommand\onedot{\futurelet\@let@token\@onedot}
\def\@onedot{\ifx\@let@token.\else.\null\fi\xspace}

\def\etal{\emph{et al}\onedot}
\makeatother

\DeclarePairedDelimiterX\set[1]\lbrace\rbrace{#1}

\title{Hard Samples Rectification for Unsupervised Cross-domain Person Re-identification}
\name{Chih-Ting Liu$^{\star}$, Man-Yu Lee$^{\star}$, Tsai-Shien Chen, Shao-Yi Chien
\thanks{${\star}$ denotes equal contribution}
\thanks{\scriptsize This research was supported in part by the Ministry of Science and Technology of Taiwan (MOST 110-2218-E-002 -025 -), National Taiwan University (NTU-108L104039), Intel Corporation, Delta Electronics and Compal Electronics.}
}
\address{Graduate Institute of Electronics Engineering, National Taiwan University, Taipei, Taiwan}
\begin{document}
%
\maketitle
\begin{abstract}
Person re-identification (re-ID) has received great success with the supervised learning methods. However, the task of unsupervised cross-domain re-ID is still challenging. In this paper, we propose a Hard Samples Rectification (HSR) learning scheme which resolves the weakness of original clustering-based methods being vulnerable to the hard positive and negative samples in the target unlabelled dataset. Our HSR contains two parts, an inter-camera mining method that helps recognize a person under different views (hard positive) and a part-based homogeneity technique that makes the model discriminate different persons but with similar appearance (hard negative). By rectifying those two hard cases, the re-ID model can learn effectively and achieve promising results on two large-scale benchmarks.
\end{abstract}
\vspace{-1mm}

\vspace{-3mm}
\vspace{-2mm}
\section{Introduction}
\label{sec:intro}
\vspace{-3mm}
Person re-identification (re-ID) tackles the problem of matching images of the same person in a camera network, which has drawn much attention in recent years because of its wide applications in the intelligent surveillance system.
Many existing works obtained great success by adopting supervised deep learning approaches~\cite{PCB,MGN}; however, it is impractical in real-world scenarios owing to the high annotation costs. Thus, how to perform re-ID in an unsupervised manner would be a critical yet challenging issue to be solved.
Cross-domain re-ID, which aims at learning re-ID on the target unlabelled domain with the aid of labelled data on a source domain, is one of the unsupervised problems that has been continuously addressed. Some works~\cite{SPGAN,HHL} utilize image-to-image translation with Generative Adversarial Network (GAN)~\cite{GAN} to translate images from source to target domain. However, those methods depend on the quality of generated fake images and ignore the inherent data distribution in the unlabelled domain.

To exploit the discriminative characteristics accessible in the target domain, the recent works mainly focus on clustering-based methods~\cite{CAMEL,PUL,PAST,SSG,eccv20} which generate pseudo identity labels by clustering the unlabelled data. Thus, the estimated correspondence of images can help for the unsupervised training.
However, an underlying main problem of clustering-based methods is that the capability of re-ID model highly relies on the ``quality'' of the clustering results. 
In other words, the inconsistency between the generated pseudo labels and the unknown ground truth labels would undesirably degrade the re-ID performance, which generally arose from the misclustered hard training pairs.
For instance, the same identity pairs captured under different cameras with intensive variations of the appearance could be possibly misclustered to different groups (we call it the hard positive). Or two people with similar appearance but only with subtle difference are likely to be clustered into the same group and be assigned with the same pseudo label (we call it the hard negative). These two situations are harmful for re-ID model learning because they all degrade the discriminative ability for identifying people. With the above observations, we propose a \textit{Hard Samples Rectification (HSR)} learning scheme which contains two components, an inter-camera mining (ICM) and a part-based homogeneity (PBH) techniques.

Because the camera ID information of each image is easily available in the dataset, based on the data feature similarity, our ICM will additionally mine and pull close those possible hard positive pairs which are mutually similar but with different camera views. This data mining is beyond the original clustering results and can steadily rectify the cluster quality afterwards.
For refining the cluster containing hard negative pairs, the proposed PBH technique will forcibly partition off and regroup the imperfect cluster with the part-based features. The idea behind is that the part-based feature gives a finer insight of a person; thus, with our PBH, the hard negative samples among a cluster will have the chance be rectified and assigned with different pseudo labels.
The main contributions of this work can be summarized as follows:
\begin{compactitem}
\item We propose an inter-camera mining technique (ICM) to mine potentially hard positive samples and alleviate the clustering bias of human appearance.
\item The proposed part-based homogeneity technique (PBH) effectively regroups the imperfect clusters containing hard negative samples. 
\item We conduct extensive experiments on two large-scale benchmarks and our HSR achieves promising performances in cross-domain unsupervised person re-ID.
\end{compactitem}

\begin{figure*}[t!]
	\centering
    \includegraphics[width=0.9\textwidth]{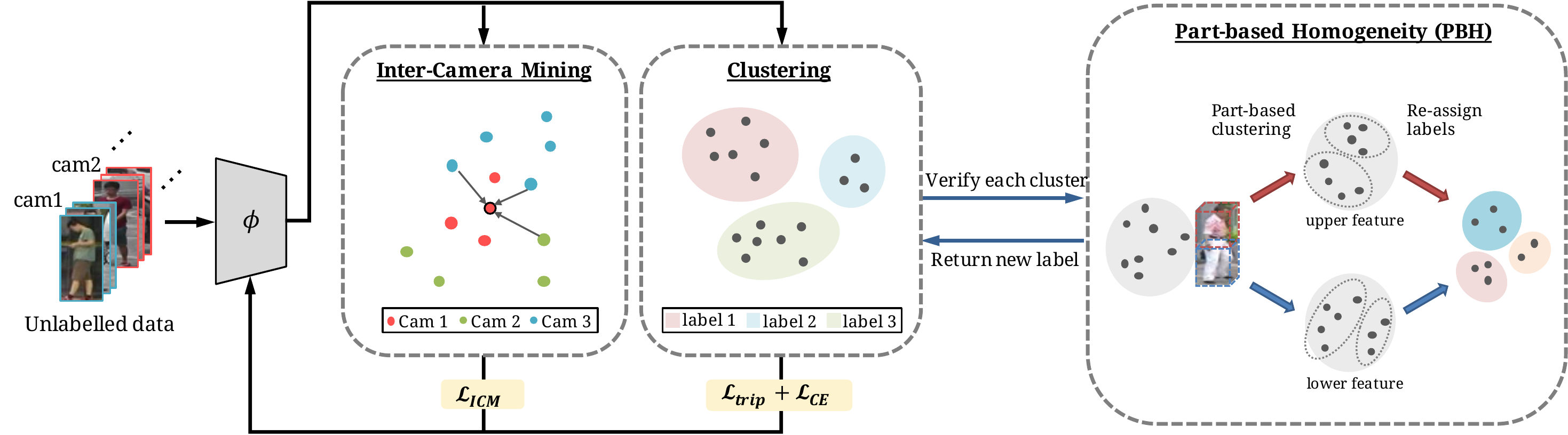}
    \mycaption{Overview of the proposed HSR learning scheme}{\small This learning scheme is conducted iteratively with clustering and network training. ICM are used to create additional hard positive training pairs and PBH is used to rectify the hard negative pairs in original clustering results. }
    \label{fig:proposed_method}
    \vspace{-6mm}
\end{figure*}

\vspace{-5mm}
\section{Proposed Methods}
\vspace{-3mm}
\subsection{Overview of our HSR learning scheme} 
\label{sub:overview}
\vspace{-2mm}
We first define the notation to be used in this paper. Given an unlabelled target dataset $\left\{ I^{t}_{c,i}\right\}^{N_t}_{i=1}$ containing total $N_t$ training images, where $c$ denotes the camera ID of image $I^t_{c,i}$, and a source labelled dataset which serves as a preliminary knowledge base for learning re-ID, the goal of our model is to learn a discriminative ability to perform person re-ID on the target dataset. 
The learning scheme is shown in Fig.~\ref{fig:proposed_method}. Typically, a feature extractor $\phi$ will be first learnt on the labelled source dataset as a pretrained feature embedding function $\phi(\cdot , {\theta}_s)$, where $\theta_s$ is the parameters learned on the source domain. 
%
Then, similar to~\cite{SSG,PAST}, a clustering algorithm called DBSCAN~\cite{DBSCAN}, which does not require the exact number of clusters or identities, will be used to generate the pseudo labels for the target unlabelled images based on the extracted feature vectors $\phi(I^{t}_{c,i} , {\theta}_s)$.
With ``estimated'' pseudo labels $y^{t}_{i}$, we can learn the re-ID model in the typical supervised manner, which consists of the cross-entropy loss ($\Lagr_{CE}$) that helps correctly classify the identities~\cite{softmax} and the triplet loss ($\Lagr_{trip}$) for controlling the distance of the positive and negative pairs in the embedding feature space~\cite{triplet}.
The clustering and network optimization stages will be conducted iteratively, and the performance of re-ID model and the quality of clustering results will improve steadily.
However, it will reach a bottleneck owing to the situations caused by the hard samples as mentioned above.
To further enhance the model ability, we propose Hard Samples Rectification (HSR) learning scheme, which dually rectifies the hard positive and negative samples with two components: inter-camera mining (ICM) and part-based homogeneity (PBH) techniques, as shown in Fig.~\ref{fig:proposed_method}.
During training, ICM will mine possible hard positive pairs with different camera views and apply triplet loss to pull close those pairs in the feature space. On the other hand, PBH technique will refine the potential imperfect clusters by splitting the hard negative pairs within the same group.

\vspace{-4mm}
\subsection{Inter-Camera Mining} 
\vspace{-2mm}
\label{sub:ICM}
As mentioned in Section~\ref{sec:intro}, hard positive pairs may be assigned to different pseudo labels due to the variance of appearance under different cameras.
After several iterations of clustering and network training, it will leads to a vicious cycle that the positive pairs used to optimize the model are only those with similar content, which goes against the goal of person re-ID to match people across cameras.
Thus, we propose an inter-camera mining technique as a role of assisting the original clustering method to mine the hard positive samples. 

\begin{algorithm}[t]
    \caption{Inter-Camera Mining}
    \begin{algorithmic}[1]
    \REQUIRE 
        Image feature vectors $\left\{\phi(I^{t}_{i})\right\}^{N_t}_{i=1}$ and its camera ID $\left\{c_i\right\}^{N_t}_{i=1}$ on target domain
    \ENSURE 
        Possible hard positive pairs
    \STATE Calculate similarity matrix $\mathbf{S}\in\mathbb{R}^{{N_t} \times {N_t}}$.
    \FOR{ $i$=$1$ ; $i\leq N_t$ ; $i$=$i$+$1$ }
    \STATE Sort $\mathbf{S}[i]$ in descending order.
    \STATE \textit{Rank}($I^{t}_{i}$) $=$ top-$K$ images $\left\{I^t_j\right\}^K_{j=1}$ in $\mathbf{S}[i]$ with $c_j \neq c_i$
    \ENDFOR
    \STATE Choose image pairs ($I^{t}_{i}$ , $I^{t}_{j}$) conform to $I^t_j \in$ \textit{Rank}($I^{t}_{i}$) and \\$I^t_i \in$ \textit{Rank}($I^{t}_{j}$).
    \STATE return all chosen pairs.
    \end{algorithmic}
    \label{alg:ICM}
\end{algorithm}

In practice, shown in Algorithm~\ref{alg:ICM}, we first compute the similarity matrix $\mathbf{S}\in\mathbb{R}^{{N_t} \times {N_t}}$ for all target images, where the element in the $i$-th row and $j$-th column is the negative Euclidean distance of  $\phi(I^{t}_{i})$ and $\phi(I^{t}_{j})$.
Then, after sorting each row in descending order, we form the possible hard positive ranking list of each image by selecting its top-$K$ closest images according to the matrix $\mathbf{S}$, denoted as \textit{Rank}($I^{t}_{i}$) with a total length of $K$. It is worth noting that in order to emphasize on ``inter-camera'' positive pairs, we remove those images captured by the camera same as the image $I^t_i$.
To ensure the robustness and correctness of our inter-camera mining, inspired by Dekel~\etal~\cite{BBP}, we additionally conduct a $K$ mutually best-buddies pairs technique. 
That is to say, for every image $I^{t}_{j}$ in \textit{Rank}($I^{t}_{i}$), $I^{t}_{i}$ should as well be in \textit{Rank}($I^{t}_{j}$). Thus, only the image pair ($I^{t}_{i}$ , $I^{t}_{j}$) that meets the above requirement would be taken into account as a reliable hard positive pair in the following CNN training.
%

%
With the mining hard positive pairs, we additionally apply the triplet loss $\Lagr_{ICM}$, where the selection of positive samples is based on our ICM mining results. Notes that it differs from the original $\Lagr_{trip}$ which samples the positives based on the pseudo labels generated by the clustering algorithm.
As for the choice of negative samples of each anchor $I^{t}_{i}$ in $\Lagr_{ICM}$, 
we choose images with different pseudo labels from $I^{t}_{i}$ and at the same time not in its rank list \textit{Rank}($I^{t}_{i}$).
%
Different from~\cite{PAST}, we embed the accessible camera information and the mutual similarity, which benefits the correctness and the robustness of additional triplet pairs mining.
With our $\Lagr_{ICM}$ iteratively shortening the distance of these mined hard positive samples, it can progressively ensure the ability of our model to match person regardless of the variation between camera views and at the same time improve the quality of the clustering results.  
\vspace{-5mm}
\subsection{Part-based Homogeneity} 
\begin{figure}[t]
	\centering
    \includegraphics[width=0.75\linewidth]{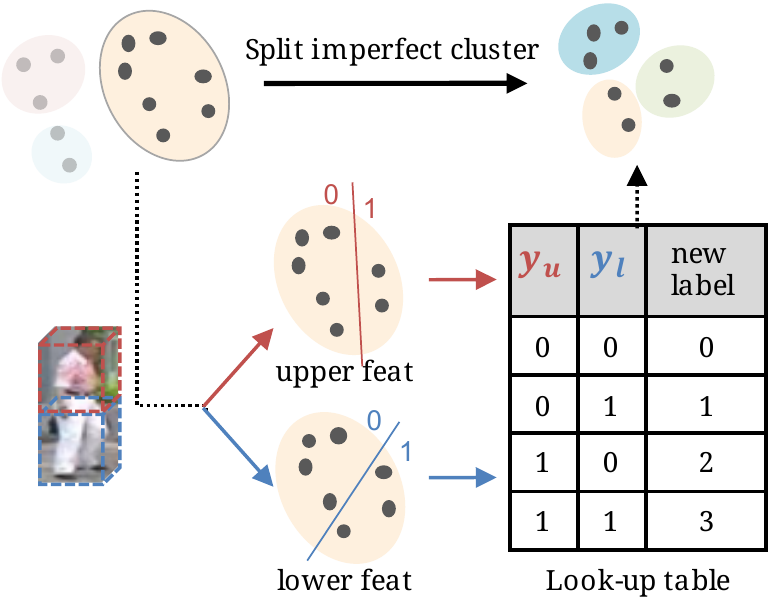}
    \mycaption{Illustration of part-based homogeneity technique}{\small We extract local features of upper part and lower part for each sample in the imperfect cluster and apply K-means clustering respectively on both local features to obtain two kinds of part-based labels. With the two temporary local labels, the cluster is then split into at most four different groups according to the look-up table.}
    \label{fig:PBS}
    \vspace{-3mm}
\end{figure}

\vspace{-2mm}
\label{sub:PBH}
Different people with only subtle difference are possibly assigned with the same pseudo labels, which would degrade the model ability to discriminatively identify people in detail.
In the aim of separating imperfect clusters which possibly contain hard negative pairs, we develop a novel method called part-based homogeneity (PBH) as a rectification technique by utilizing the local features which provide finer information other than the global one.
First, we need to define the imperfectness of a cluster and select the candidates for applying our PBH. To this end, we utilize Silhouette score~\cite{silhouette}, which is an evaluation metric for measuring how well a sample is clustered to its group without the requirement of the ground truth labels. 
By computing the mean Silhouette score of data in each cluster $i$, denoted as $mSil(i)$, we can further select the imperfect cluster with its $mSil(i)$ smaller than an empirically predefined threshold $\lambda$.
Our proposed PBH technique is then applied on every selected cluster to refine the original clustering results, as illustrated in Fig.~\ref{fig:PBS}.

To start with, we split and pool the output feature maps of every sample in the selected cluster $j$ into two parts: upper and lower features, which are formulated as $\left\{ f_{u,i}\right\}^{N_j}_{i=1}$ and $\left\{ f_{l,i}\right\}^{N_j}_{i=1}$, where $N_j$ is the number of samples in cluster $j$.
Then, we respectively employ the K-means clustering with K = 2 on $\left\{ f_{u,i}\right\}^{N_j}_{i=1}$ and $\left\{ f_{l,i}\right\}^{N_j}_{i=1}$ to observe the data distribution of the finer local features.
%
Consequently, each sample is assigned with two temporary labels based on the groups of its upper and lower features, denoted as $y_u$ and $y_l$. 
With the part-based label pair $(y_u, y_l)$, we can re-assign new pseudo labels to the samples in cluster $j$ according to a look-up table, as shown in Fig.~\ref{fig:PBS}.
The idea behind is that only the data with both similar local parts, which means the same $(y_u, y_l)$, can be assigned with the same pseudo label.
Notably, because the number of contained ground truth labels is unknown, we suppose that if the cluster is defined as an imperfect one, it would contain at least two ground truth labels. Furthermore, the progress of iterative learning can ensure that even the selected imperfect cluster contains more than two ground truth labels, the split clusters would still have the chance to be defined as imperfect ones in the next iteration. 
In summary, by considering the local features, our PBH maintains the homogeneity within the new cluster and avoids assigning the same pseudo label to globally similar hard negative pairs.   

\vspace{-4mm}
\subsection{Optimization Procedure} 
\vspace{-2mm}
\label{sub:optimization}
For each iteration, after clustering the unlabelled data by DBSCAN, we would first verify the imperfect clusters and adopt the proposed PBH technique to refine the original estimated pseudo labels. Then, we jointly utilize the triplet loss ($\Lagr_{trip}$) and the cross-entropy loss ($\Lagr_{CE}$) to optimize the CNN network with those updated pseudo labels. Besides the positive and negative pairs sampled from the pseudo labels, we also jointly adopted the triplet loss $\Lagr_{ICM}$ according to our ICM sampling technique.
The overall loss function can be written as follows:
\vspace{-4mm}
\begin{equation}
\vspace{-3mm}
\label{eq:total}
\Lagr_{total} = \Lagr_{CE} + \Lagr_{trip} + \Lagr_{ICM}
\end{equation}
%
%
%
%

\section{Experiments}
\label{experiments}
\vspace{-3mm}
\subsection{Datasets and Evaluation Protocol}
\vspace{-2mm}
We evaluate our approach on two large-scale person re-ID benchmarks: Market-1501~\cite{Market1501} and DukeMTMC-ReID~\cite{DukeReID}, abbreviated as ``Market'' and ``Duke'' in the following sections.
The Market and Duke datasets contain 1501 and 1404 identities respectively, and each of the identity was captured by at most 6 or 8 cameras.
In our experiments, the label information of the training data in the target domain is not available during the whole learning process. Rank-1 (R1) accuracy (\%) and the Mean Average Precision (mAP,~\%) are used to evaluate the re-ID performance.
\vspace{-3mm}
\subsection{Implementation Details}
\vspace{-2mm}
We adopt ResNet-50~\cite{resnet} as our feature extractor $\phi$ and use the last 2048-d feature vector to represent the data in both training and clustering. 
Notes that we split the last feature map before average pooling into the upper and lower local feature parts in our proposed PBH technique. 
The image size is $256\times 128$ and augmented with random erasing and horizontal flip. Each mini-batch is with size 32, which consists of 8 randomly sampled pseudo identities, and for $\Lagr_{trip}$, each contains 4 sampled images in their cluster, but for $\Lagr_{ICM}$, the 4 samples come from the possible hard positive ranking list.
Empirically, we set $K$ = 10 in the ICM, the number of local features = 2 (upper and lower) in PBH, and set the threshold $\lambda = mean(mSil)-3std(mSil)$ in our PBH, where $mean(mSil)$ and $std(mSil)$ denote the average and standard deviation of $mSil$ of all clusters. 
We choose the SGD optimizer with the learning rate = 0.005 to optimize the model for 10 epochs in each iteration, where the total \#iterations is 30. 
\vspace{-3mm}
\subsection{Comparison with State-of-the-arts}
\vspace{-2mm}
We compare our proposed HSR with existing state-of-the-art unsupervised cross-domain re-ID methods on Market and Duke datasets in Table~\ref{sub:SOTA}. Based on the common settings, we use Duke as the source dataset when test on Market and vice versa.
We can see that our HSR outperforms all the compared methods significantly on both datasets.
\begin{table}[t]
\centering
\caption{Comparisons with state-of-the-arts unsupervised re-ID methods on Market and Duke.}
\scalebox{0.9}{
\begin{tabular}{l|cc|cc}
\hline
\multirow{2}{*}{Methods} & \multicolumn{2}{c|}{Duke $\rightarrow$ Market} & \multicolumn{2}{c}{Market $\rightarrow$ Duke} \\ 
\cline{2-5} 
    & R1 & mAP & R1 & mAP \\ \hline\hline
PUL~\cite{PUL} & 45.5  & 20.5 & 30.0  & 16.4 \\
CAMEL~\cite{CAMEL} & 54.5 & 26.3 & - & - \\
SPGAN~\cite{SPGAN} & 58.1 & 26.9 & 46.9 & 26.4 \\
HHL~\cite{HHL} & 62.2 & 31.4 & 46.9 & 27.2 \\
MAR~\cite{MAR} & 67.7 & 40.0 & 67.1 & 48.0 \\
PAST~\cite{PAST} & 78.4 & 54.6 & 72.4 & 54.3 \\
SSG~\cite{SSG} & 80.0 & 58.3 & 73.0 & 53.4 \\ 
\textit{p}MR-SADA~\cite{SADA}& 83.0 & 59.8 & 74.5 & 55.8 \\
GDS-H~\cite{eccv20} & 81.1 & 61.2 & 73.1 & 55.1 \\ 
\hline
\textbf{HSR (Ours)} & \textbf{85.3} & \textbf{65.2} & \textbf{76.1} & \textbf{58.1} \\ \hline
\end{tabular}}
\vspace{-3mm}
\label{sub:SOTA}
\end{table}
Among the compared methods, the works~\cite{PUL,CAMEL} and some latest approaches~\cite{PAST,SSG,eccv20} also aim to exploit discriminative information in target domain based on pseudo label estimation. Different from them, our HSR focuses on mining hard positive and hard negative samples to calibrate the unreliable clustering results, and thus acquires a promising gain in the performance.
Specifically, HSR achieves R1 = $85.3\%$ and mAP = $65.2\%$, which outperforms the best of these approaches by margins of $\mathbf{4.2}\%$ and $\mathbf{4.0}\%$ in Market.
Similar improvement can be seen in Duke by achieving R-1 = $76.1\%$ and mAP = $58.0\%$, with  margins of $\mathbf{3.0}\%$ and $\mathbf{3.0}\%$.
In summary, our method effectively enhances the model capability by alleviating the effect of hard cases in clustering-based methods.
\begin{table}[t!]
\centering
\caption{Ablation studies of the proposed methods in terms of R1 and mAP (\%).}
\scalebox{0.9}{
\begin{tabular}{l|cc|cc}
\hline
\multirow{2}{*}{Experimental setting} & \multicolumn{2}{c|}{Duke $\rightarrow$ Martket} & \multicolumn{2}{c}{Market $\rightarrow$ Duke} \\ \cline{2-5} 
 &  R1 & mAP & R1 & mAP \\ \hline \hline
Direct Transfer & 50.1 & 20.9 & 36.2 & 18.3 \\
Baseline &  72.9 & 46.3 & 60.2 & 42.2 \\
Baseline + PBH  & 74.5 & 47.1 & 63.5 & 44.6 \\
Baseline + $\Lagr_{ICM}$ & 83.8 & 63.3 & 73.5 & 54.4 \\ \hline
\textbf{HSR (Ours)} &  \textbf{85.3} & \textbf{65.2} & \textbf{76.1} & \textbf{58.0} \\ \hline
\end{tabular}}
\label{tab:ablation}
\vspace{-4mm}
\end{table}

\vspace{-3mm}
\subsection{Ablation Study}
\vspace{-2mm}
We perform ablation study to evaluate the effectiveness of each proposed component. Results are shown in Table~\ref{tab:ablation}. First, we directly apply the model pretrained on source dataset to the target dataset, denoted as ``Direct Transfer''. The inferior performance due to the discrepancy between domains reveals the necessity of applying unsupervised method on target domain. Next, for the ``Baseline'' method, we utilize DBSCAN to cluster and generate pseudo labels for learning re-ID model on target domain, as shown in the second row of the Table~\ref{tab:ablation}; the baseline can achieve $46.3\%$ and $42.2\%$ in terms of mAP on the two datasets respectively. But the unsatisfactory results can be explained by the inaccurate pseudo labels, thus degrades the model capability to identify persons. \\
\textbf{Effectiveness of PBH and ICM.}~~The hard negative pairs being clustered to a same group is a critical factor that hinders the model ability to distinguish different identities in details. With our PBH, the imperfect clusters will be split into multiple groups and re-assigned new pseudo labels. The third row of Table~\ref{tab:ablation} (Baseline + PBH) shows that with PBH, all performance results improve on the two datasets compared to the baseline method.
We then validate the proposed ICM in the fourth row of Table~\ref{tab:ablation} (Baseline + $\Lagr_{ICM}$). A significant improvement can be observed compared to baseline, which gains $\mathbf{17.0}\%$ and $\mathbf{12.2}\%$ in mAP on Market and Duke. This demonstrates that our inter-camera mining is able to assist the model learning to identify people regardless of the cross-camera scene variation, which is exactly the goal of re-ID.
We also calculate the precision of the selected positive pairs from ICM.
Specifically, for every sample $I^{t}_{i}$ in the target domain, we compute the average of true positive rate of their corresponding \textit{Rank}($I^{t}_{i}$). It can progressively rise up to $82\%$ in our iterative learning process, and notably, we can also reach a maximum rate of $17\%$ in our \textit{Rank}($I^{t}_{i}$) of ``hard positive samples'', which possesses the same ground truth identities yet are assigned into ``different'' clusters. This shows that our ICM has strong capability of rectifying the original clustering results and favorably generate possible hard positive for model learning.
%

\vspace{-5mm}
\section{Conclusion}
\vspace{-3mm}
\label{sub:conclusion}
In this paper, we introduce a hard samples rectification (HSR) learning scheme to address the issue of hard samples that degrades the performance in clustering-based methods. Specifically, we propose an inter-camera mining technique to match people under various camera views, and a part-based homogeneity technique to split hard negative pair within same cluster in a part-based manner. With our HSR, the model can learn a discriminative representation for unlabelled target images and receive a significant improvement of re-ID performance.

\bibliographystyle{IEEEbib}
\bibliography{egbib}

\end{document}